# CNN-LTE: A CLASS OF 1-X POOLING CONVOLUTIONAL NEURAL NETWORKS ON LABEL TREE EMBEDDINGS FOR AUDIO SCENE RECOGNITION

*Huy Phan*[⋆†], *Lars Hertel*[⋆], *Marco Maass*[⋆], *Philipp Koch*[⋆], and *Alfred Mertins*[⋆]

[⋆]Institute for Signal Processing, University of Lübeck, Germany
[†]Graduate School for Computing in Medicine and Life Sciences, University of Lübeck, Germany
Email: {phan, hertel, maass, koch, mertins}@isip.uni-luebeck.de

## ABSTRACT

We describe in this report our audio scene recognition system submitted to the DCASE 2016 challenge [1]. Firstly, given the label set of the scenes, a label tree is automatically constructed. This category taxonomy is then used in the feature extraction step in which an audio scene instance is represented by a label tree embedding image. Different convolutional neural networks, which are tailored for the task at hand, are finally learned on top of the image features for scene recognition. Our system reaches an overall recognition accuracy of 81.2% and 83.3% and outperforms the DCASE 2016 baseline with absolute improvements of 8.7% and 6.1% on the development and test data, respectively.

## 1. INTRODUCTION

Acoustic scene classification (ASC) is an important problem of computational auditory scene analysis [2, 3]. Solving this problem will allow a device to recognize a surrounding environment via the sound it captures, and hence, enables a wide range of applications, such as surveillance [4], robotic navigation [5], and context-aware services [6, 7]. However, due to its complex sound composition, it is challenging to obtain a good representation for recognition.

In our proposed system, we firstly obtain label tree embedding (LTE) representations for audio scenes as in [8]. Specifically, a class taxonomy is constructed by learning to group similar categories into meta-classes on a tree structure. We then derive explicit embeddings to map each audio segment into the semantic space that underlies the class hierarchy. However, instead of producing a global feature vector for a scene instance by average pooling [8], we represent it with the 2-dimensional LTE image. Afterward, we trained different 1-X pooling convolutional neural networks (CNN) [9], including *1-max*, *1-mean*, and *1-mix pooling CNNs*, on top of these images for recognition. While the first network is expected to uncover most discriminative foreground events of the scenes, the second one tends to capture the average background, and the third one is to combine both types of information into the same model.

## 2. RECOGNITION WITH LTE REPRESENTATIONS

### 2.1. LTE representations for audio scenes

#### 2.1.1. Learning a label tree

Consider a database (e.g. scene database) with the label set $\mathcal{L} = \{1, \ldots, C\}$ where $C$ indicates the number of target categories. In order to explore the structure of class labels, we learn a label tree similar to [10, 11]. The learning algorithm collectively partitions the label set into disjoint subsets in such a way that they are easy to distinguish from one another. Given the set of samples $\mathcal{S} = \{(\mathbf{x}_n, c_n)\}_{n=1}^{|\mathcal{S}|}$ extracted from the training data, where $\mathbf{x} \in \mathbb{R}^M$ denotes the vector of some $M$ low-level features, $c \in \mathcal{L}$ indicates the class label, and $|\cdot|$ represents the set cardinality.

The label tree is constructed recursively so that each node is associated with a set of class labels. Consider a node with a label set $\ell$ (and therefore, the root node is assigned with the label set $\mathcal{L}$), our goal is to split $\ell$ into two subsets $\ell^L$ and $\ell^R$ that hold the following requirements: $\ell^L \neq \emptyset$, $\ell^R \neq \emptyset$, $\ell^L \cup \ell^R = \ell$, and $\ell^L \cap \ell^R = \emptyset$. There are totally $2^{|\ell|-1} - 1$ such possible partitions $\{\ell^L, \ell^R\}$. The optimal partition is then adopted such that a binary classifier designed to separate $\ell^L$ and $\ell^R$ makes as few errors as possible.

In order to find the optimal partitioning, we rely on the multi-class confusion matrix which indicates how good a class is separated from the others. Let $\mathcal{S}^\ell \subset \mathcal{S}$ denote the set of samples corresponding to the label set $\ell$. Furthermore, suppose that we have changed and sorted the label set $\ell$ so that $\ell = \{1, \ldots, |\ell|\}$. In addition, we divide $\mathcal{S}^\ell$ into two equal halves: $\mathcal{S}^\ell_{\text{train}}$ for training a classifier and $\mathcal{S}^\ell_{\text{eval}}$ for evaluation. We train the multi-class classifier $\mathcal{M}^\ell$ using random forest classification [12] with 200 trees using $\mathcal{S}^\ell_{\text{train}}$ and then evaluate it on the evaluation set $\mathcal{S}^\ell_{\text{eval}}$ to obtain the confusion matrix $\mathbf{A} \in \mathbb{R}^{|\ell| \times |\ell|}$. Each element $\mathbf{A}_{ij}$ of the matrix $\mathbf{A}$ is computed by

$$\mathbf{A}_{ij} = \frac{1}{|\mathcal{S}^\ell_{\text{eval},i}|} \sum_{\mathbf{x} \in \mathcal{S}^\ell_{\text{eval},i}} P(j|\mathbf{x}, \mathcal{M}^\ell). \quad (1)$$

Here, $\mathcal{S}^\ell_{\text{eval},i} \subset \mathcal{S}^\ell_{\text{eval}}$ is the set of samples with the label $i$. $P(j|\mathbf{x}, \mathcal{M}^\ell)$ denotes the probability that the classifier $\mathcal{M}^\ell$ predicts the sample $\mathbf{x}$ as class $j$. $\mathbf{A}_{ij}$ implies how likely a sample of the class $i$ is wrongly predicted to belong to the class $j$ by the classifier. Since $\mathbf{A}$ is not symmetric, we symmetrize it as

$$\bar{\mathbf{A}} = (\mathbf{A} + \mathbf{A}^\mathsf{T})/2. \quad (2)$$

Eventually, the optimal partitioning $\{\ell^L, \ell^R\}$ is selected to maximize:

$$E(\ell) = \sum_{i,j \in \ell^L} \bar{\mathbf{A}}_{ij} + \sum_{m,n \in \ell^R} \bar{\mathbf{A}}_{mn}. \quad (3)$$

By this, we tend to group the ambiguous classes into the same subset, as a result, produce two meta-classes $\{\ell^L, \ell^R\}$ that are easy to separate from each other. We apply spectral clustering [13] on the matrix $\bar{\mathbf{A}}$ to solve a relaxed version of the optimization problem in (3). The subsets $\ell^L$ and $\ell^R$ are then directed to the left and right child nodes, respectively. The splitting process is recursively repeated to grow the whole tree until a leaf node with a single class label is reached.

*2.1.2. Label tree embedding representations*

Via the learned label tree, we have formed $(C-1) \times 2$ meta-classes in total from the original label set $\mathcal{L}$. Two of them are associated with the left and right child nodes of one out of $(C-1)$ split nodes. For clarity, suppose that we have indexed the split nodes of the label tree as $\{\ell_i\}_{i=1}^{C-1}$. Our objective is then to learn a representation for a test sample by embedding them into the space of the meta-class labels. Formally, we then want to obtain an explicit mapping $\Psi : \mathbb{R}^M \to \mathbb{R}^{(C-1) \times 2}$ to map the test sample $\mathbf{x} \in \mathbb{R}^M$ to a feature vector $\Psi(\mathbf{x}) = (\psi_1^L(\mathbf{x}), \psi_1^R(\mathbf{x}), \ldots, \psi_{C-1}^L(\mathbf{x}), \psi_{C-1}^R(\mathbf{x}))$. The entries of $\psi_i^L(\mathbf{x})$ and $\psi_i^R(\mathbf{x})$ denote the likelihoods that $\mathbf{x}$ belongs to two meta-classes on the left and right child nodes of the split node $\ell_i$.

To obtain the likelihoods, at a split node $\ell_i$ with the optimal partition $\{\ell_i^L, \ell_i^R\}$, we train the binary random-forest classifier $\mathcal{M}_i^\ell$ with 200 trees using the whole set $S^{\ell_i}$ as training data. The samples with their labels in $\ell_i^L$ are considered as negative examples and others with their labels in $\ell_i^R$ are considered as positive examples. The likelihoods are then given by:

$$\psi_i^L(\mathbf{x}) = P(\text{negative}|\mathbf{x}, \mathcal{M}^{\ell_i}), \quad (4)$$
$$\psi_i^R(\mathbf{x}) = P(\text{positive}|\mathbf{x}, \mathcal{M}^{\ell_i}). \quad (5)$$

Here, $P(\text{negative}|\mathbf{x}, \mathcal{M}^{\ell_i})$ and $P(\text{positive}|\mathbf{x}, \mathcal{M}^{\ell_i})$ are the classification probabilities outputted by $\mathcal{M}^{\ell_i}$ when evaluating on $\mathbf{x}$, thanks to the probability support of the random forest classification [12].

## 2.2. Recognition using LTE representations

The LTE representations can be used for scene recognition as in [8]. Using the above framework, we derived the following LTE representations with different low-level features: (1) Gammatone cepstral coefficients (LTE1), (2) MFCCs (LTE2), log frequency filter banks (LTE3), and their combination altogether (LTE+).

**LTE1**. We extract $M = 64$ Gammatone ceptral coefficients [14] in the frequency range of 20-22050 Hz. It should be noticed that we do not consider the whole 30-second snippet of an acoustic scene instance as a sample. Instead, in order to capture meaningful events happening in a scene whose lengths are in order of hundreds of milliseconds, we use segments of length 500 ms with an overlap of 250 ms as the samples for further processing. This results in $T = 118$ segments for each 30-second snippet. Each segment is then decomposed into 50 ms frames with 50% overlap, each of which is represented by a 64-dimensional feature vector computed by averaging the feature vectors of its constituent frames. Furthermore, each audio segment is labeled by the label of the scene where it stemmed from. This eventually resulted in a $F \times T$ LTE image for each 30-second scene instance where $F = (C-1) \times 2$. In order to train classifier based on these LTE features, we performed *average pooling* on a $F \times T$ LTE image over time to obtain the global $F$-dimensional feature vector for each scene instance.

**LTE2**. For these LTE features, we employed $M = 60$ MFCC features (including 20 MFCC static coefficients, 20 delta MFCC coefficients, and 20 acceleration MFCC coefficients) as in the baseline in replacement for Gammatone ceptral coefficients in the LTE1.

**LTE3**. We utilized the set of features in our previous works [15, 16, 11, 10] as low-level features in replacement for Gammatone ceptral coefficients in the LTE1. They include 20 log-frequency filter bank coefficients, their first and second derivatives, zero-crossing rate, short-time energy, four sub-band energies, spectral centroid, and spectral bandwidth. The overall feature dimension is $M = 65$.

**LTE+**. In order to take advantage of representations from different perspectives (i.e. different low-level features), we combine them using the extended Gaussian-$\chi^2$ kernel [17] given by

$$K(\mathbf{x}_i, \mathbf{x}_j) = \exp\Big(-\sum_k \frac{1}{\bar{D}^k} D\big(\Psi^k(\mathbf{x}_i), \Psi^k(\mathbf{x}_j)\big)\Big) \quad (6)$$

where $D\big(\Psi^k(\mathbf{x}_i), \Psi^k(\mathbf{x}_j)\big)$ is the $\chi^2$ distance between the embedded scene instances $\Psi^k(\mathbf{x}_i)$ and $\Psi^k(\mathbf{x}_j)$ with respect to the $k$-th channel where $k \in \{\text{LTE1}, \text{LTE2}, \text{LTE3}\}$. $\bar{D}^k$ is the mean $\chi^2$ distance of the embedded scene instances in training data for the $k$-th channel.

To extract descriptors for the training instances, we conducted 10-fold cross-validation on training data. We trained the final scene classification systems using one-vs-one support vector machines (SVM) with $\chi^2$ kernel. For LTE+, we used nonlinear SVMs with the extended Gaussian kernel given in (6). The hyperparameters of the SVMs were tuned via 10-fold cross-validation.

## 2.3. Potential issues

We have shown in our previous work [8] that this recognition scheme achieves state-of-the-art performance on different audio scene datasets [8], thanks to the discriminative powers of LTE features. However, we argue that the average pooling on the $F \times T$ images resutls in global feature vectors that are not optimal.

Excluding the background noise, an acoustic scene usually involves various kinds of foreground sounds. As a result, it can be interpreted as foreground events on the bed of background noise. Foreground events [18, 19, 20] and background noise [21] have been used as a footprint to represent a scene. However, they should be considered separately [22]. Unfortunately, with the average pooling, we tend to mix up the foreground events and background noise. To overcome this issue, we propose a classification scheme using 1-X pooling CNNs on the LTE images where 1-X pooling stands for 1-max, 1-mean, and 1-mix pooling operators.

## 3. RECOGNITION WITH 1-X POOLING CNNS ON MULTI-CHANNEL LTE IMAGES

The proposed network architecture consists of three layers, including convolutional, pooling, and softmax layers [9]. We illustrate in Figure 1 one CNN with the 1-max pooling operator.

### 3.1. Multi-channel LTE images

The inputs to the networks are the whole LTE images. Furthermore, the recognition results on the development data reveal that different low-level features (e.g. Gammatone sceptral coefficients, MFCCs, and log-frequency filter banks) used to derive LTE images are good for different scene categories. In addition, background noise is shown useful for some categories but not for others. Therefore, we additionally produce three more LTE images for each scene instance by reprocessing the signal with background noise subtraction [23]. We finally stack the individual LTE images to produce the multi-channel LTE image of size $P \times F \times T$ for the scene instance when $P = 6$ is the number of single LTE images.

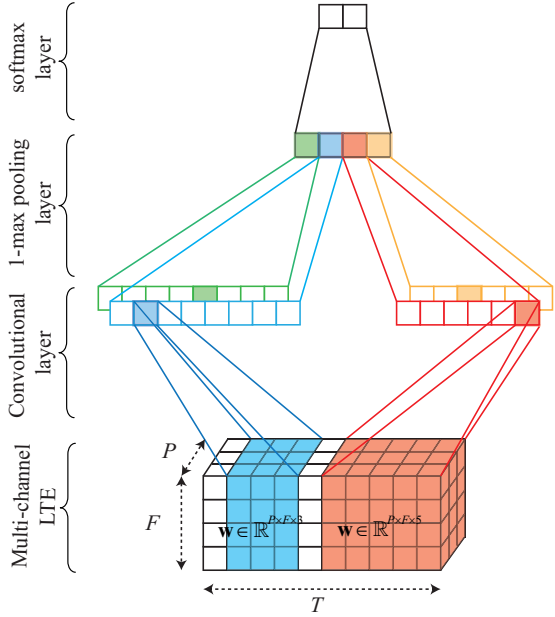

Figure 1: Illustration of 1-max pooling CNN architecture on a $P$-channel LTE image. The network consists of two filter sets with two different widths $w = \{3, 5\}$ at the convolutional layer. There are two individual filters on each filter set.

### 3.2. Convolutional layer

We aim to use the convolutional layer to extract discriminative features within the whole signals that are useful for the classification task at hand. Suppose that an LTE image presented to the network is given in the form of a 3-dimensional matrix $\mathbf{S} \in \mathbb{R}^{P \times F \times T}$. We then perform convolution on it via 3-dimensional linear filters. For simplicity, we only consider convolution in time direction, i.e. fix two dimensions of a filter to be equal to $P$ and $F$ and vary the remaining dimension to cover different number of adjacent audio segments.

Let us denote a filter by the weight matrix $\mathbf{w} \in \mathbb{R}^{P \times F \times w}$ with the width of $w$ audio segments. Therefore, the filter contains $P \times F \times w$ parameters that need to be learned. We further denote the temporal adjacent spectral slices (e.g. audio segments) from $i$ to $j$ by $\mathbf{S}[i:j]$. The convolution operation $*$ between $\mathbf{S}$ and $\mathbf{w}$ results in the output vector $\mathbf{O} = (o_1, \ldots, o_{T-w+1})$ where:

$$o_i = (\mathbf{S} * \mathbf{w})_i = \sum_{k,l,m} (\mathbf{S}[i:i+w-1] \odot \mathbf{w})_{k,l,m}. \quad (7)$$

Here, $\odot$ denotes the element-wise multiplication. We then apply an activation function $h$ to each $o_i$ to induce the feature map $\mathbf{A} = (a_1, \ldots, a_{T-w+1})$ for this filter:

$$a_i = h(o_i + b), \quad (8)$$

where $b \in \mathbb{R}$ is a bias term. Among the common activation fuctions, we chose *Rectified Linear Units* (ReLU) [24] due to their computational efficiency:

$$h(x) = \max(0, x). \quad (9)$$

To allow the network to extract complementary features and enrich the representation, we learn $Q$ different filters simultaneously. Moreover, foreground events in a scene may have different durations. We learn filters with different sizes simultaneously in order to capture them more efficiently. More specifically, we learn $R$ different sets of $Q$ filters, each of which has different width $w$ to form $Q \times R$ filters in total.

### 3.3. 1-X pooling layer

The feature maps produced by the convolution layer are forwarded to the pooling layer. We propose three different pooling operations that are especially designed for scene recognition. In addition, this pooling strategy offers a unique advantage. That is, although the dimensionality of the feature maps varies depending on the length of audio events and the width of the filters, the pooled feature vectors have the same size [9, 25, 26]. Therefore, the signals can be of any arbitrary size instead of being fixed to 30-second long as the common setting for the task.

**1-max pooling**. This pooling operation on a feature map aims to reduce a feature set to a single most dominant feature [27]. Coupled with the 1-max pooling function, each filter in the convolutional layer is optimized to detect a specific event that is allowed to occur at any time in a scene signal. Pooling on $Q \times R$ feature maps results in $Q \times R$ features that will be joined to form a feature vector that is then presented to the final softmax layer.

**1-mean pooling**. A feature map is averaged to result in a single mean feature. Due to averaging, this feature is supposed to capture the average background of the signal. $Q \times R$ features are produced from $Q \times R$ feature maps.

**1-mix pooling**. This operation performs both 1-max and 1-mean pooling at the time to encode both foreground events and the average background. The final feature vector contains $2 \times Q \times R$ features, one half consists of 1-max features and other half of 1-mean features.

### 3.4. Softmax layer

The fixed-size feature vector after the pooling layer is subsequently presented to the standard softmax layer to compute the predicted probability over the class labels. The network is trained by minimizing the cross-entropy error. This is equivalent to minimizing the KL-divergence between the prediction distribution $\hat{y}$ and the target distribution $y$. With the binary one-hot coding scheme and the network parameter $\Theta$, the error for $N$ training samples is given by

$$E(\Theta) = -\frac{1}{N} \sum_{i=1}^{N} y_i \log(\hat{y}_i(\Theta)) + \frac{\lambda}{2} ||\Theta||^2. \quad (10)$$

The hyper-parameter $\lambda$ governs the trade-off between the error term and the $\ell_2$-norm regularization term. Furthermore, we also employ dropout [28] at this layer by randomly setting values in the weight vector to zero with a predefined probability. The optimization is performed using the *Adam* gradient descent algorithm [29].

## 4. EXPERIMENTS

### 4.1. Experimental setup

The setup is based on the development setting as described in Task 1 of the DCASE 2016 challenge [1, 30]. The signals were recorded with a sampling frequency of 44.1 kHz. The development data consists of 30-second audio signals of 15 scene classes divided into 4-fold cross-validation. The average classification accuracy over all folds is reported. Especially, to handle the errors in some of the recordings, we simply remove erroneous segments from the signals. This however resulted in an LTE image with less than $T = 118$ in time which was then circularly padded to make it 118 segments long in time dimension.

Table 1: Audio scene recognition accuracy (%).

| Category | Development set | | | | | | | | | | | | Test set | |
|---|---|---|---|---|---|---|---|---|---|---|---|---|---|---|
| | Baseline | *w/ background noise* | | | | *w/o background noise* | | | | 1-Max CNN-LTE | 1-Mean CNN-LTE | 1-Mix CNN-LTE | Baseline | 1-Mix CNN-LTE |
| | | LTE1 | LTE2 | LTE3 | LTE+ | LTE1 | LTE2 | LTE3 | LTE+ | | | | | |
| Beach | 69.3 | 83.3 | 79.5 | 80.8 | 78.2 | 69.2 | 73.1 | 69.2 | 73.1 | 82.1 | 82.1 | 87.2 | 84.6 | 84.6 |
| Bus | 79.6 | 74.4 | 59.0 | 51.3 | 61.5 | 70.5 | 46.2 | 48.7 | 50.0 | 71.8 | 74.4 | 75.6 | 88.5 | 96.2 |
| Cafe/Rest. | 83.2 | 57.7 | 67.9 | 60.3 | 71.8 | 30.7 | 70.5 | 60.3 | 60.3 | 71.8 | 62.8 | 69.2 | 69.2 | 53.8 |
| Car | 87.2 | 69.2 | 65.4 | 79.5 | 79.5 | 74.4 | 80.8 | 87.2 | 88.5 | 91.0 | 91.0 | 92.3 | 96.2 | 100.0 |
| City center | 85.5 | 84.6 | 88.5 | 92.3 | 89.7 | 82.1 | 97.4 | 96.2 | 91.0 | 92.3 | 89.7 | 91.0 | 80.8 | 100.0 |
| Forest path | 81.0 | 74.0 | 80.5 | 85.7 | 85.7 | 80.5 | 74.0 | 85.7 | 88.3 | 87.0 | 88.3 | 87.0 | 65.4 | 96.2 |
| Grocery store | 65.0 | 80.8 | 92.3 | 88.5 | 87.2 | 82.1 | 89.7 | 85.9 | 87.2 | 93.6 | 87.2 | 87.2 | 88.5 | 84.6 |
| Home | 82.1 | 88.5 | 82.1 | 92.3 | 88.5 | 92.3 | 80.8 | 92.3 | 91.0 | 91.0 | 91.0 | 92.3 | 92.3 | 88.5 |
| Library | 50.4 | 82.1 | 62.8 | 76.9 | 80.8 | 51.3 | 62.8 | 73.1 | 67.9 | 79.5 | 79.5 | 80.8 | 26.9 | 46.2 |
| Metro station | 94.7 | 80.0 | 81.3 | 78.7 | 84.0 | 86.7 | 86.7 | 81.3 | 86.7 | 92.0 | 90.7 | 93.3 | 100.0 | 84.6 |
| Office | 98.6 | 96.2 | 92.3 | 85.9 | 93.6 | 100.0 | 74.4 | 84.6 | 89.7 | 98.7 | 98.7 | 100.0 | 96.2 | 100.0 |
| Park | 13.9 | 32.1 | 23.1 | 26.9 | 30.8 | 26.9 | 42.3 | 33.3 | 30.8 | 39.7 | 50.0 | 47.4 | 53.8 | 88.5 |
| Res. area | 77.7 | 71.8 | 62.8 | 69.2 | 65.4 | 69.3 | 76.9 | 76.9 | 79.5 | 76.9 | 74.4 | 66.7 | 88.5 | 84.6 |
| Train | 33.6 | 29.5 | 34.6 | 46.1 | 38.5 | 48.7 | 46.2 | 42.3 | 60.3 | 48.7 | 52.6 | 57.7 | 30.8 | 46.2 |
| Tram | 85.4 | 84.6 | 76.9 | 76.9 | 84.6 | 96.2 | 80.8 | 85.9 | 88.5 | 88.5 | 84.6 | 89.7 | 96.2 | 96.2 |
| **Overall** | 72.5 | 72.6 | 69.9 | 72.8 | 74.7 | 70.7 | 72.2 | 73.5 | 75.5 | 80.3 | 79.8 | 81.2 | 77.2 | 83.3 |

Table 2: Hyper-parameters of the proposed CNN networks.

| Hyper-parameter | Value |
|---|---|
| Filter width $w$ | $\{3, 5, 7\}$ |
| Number of filter $P$ for each size | 1000 |
| Learning rate for the Adam optimizer | 0.0001 |
| Dropout rate | 0.5 |
| Regularization parameter $\lambda$ | 0.001 |

The proposed 1-X pooling CNNs involve different hyperparameters which are specified in Table 1. The filter width $w$ is set to $\{3, 5, 7\}$ segments which are equivalent to 1, 1.5, and 2 seconds duration. The networks were trained for 500 epochs with a minibatch size of 50. In fact, the training history shows that the training converged very fast after a few dozens of epochs, and the networks do not experience overfitting after convergence.

The provided baseline system [30] consists of 60 MFCC features and a Gaussian mixture model (GMM) based classifier. MFCCs were calculated using 40 ms frames with Hamming window and 50% overlap and 40 mel bands. They include the first 20 coefficients (including the 0th order coefficient) and delta and acceleration coefficients calculated using a window length of 9 frames. A GMM model with 32 components was trained for each scene class.

### 4.2. Experimental results

The audio scene recognition accuracies obtained by our different proposed systems as well as the baseline are presented in Table 1.

On the development set, the performance of the systems based on individual LTE features is comparable with that of the baseline. When combining them in the LTE+ system, the accuracy is boosted about 2% and surpasses the baseline with 3% absolute improvement. As expected, the performance is significantly improved with the 1-X pooling CNNs, reaching 80.3%, 79.8%, and 81.2% with 1-max, 1-mean, and 1-mix pooling schemes. The result with the 1-mix CNN-LTE outperforms that of the baseline by 8.7% absolute.

### 4.3. The final submission system and results

Our final submission is based on the 1-mix pooling CNN which obtained the best performance on the development data. The whole development data is used for training purpose. Multi-channel LTE images were first generated as described in Section 3.1, and then presented to the network for training. The network was trained for 100 epochs with a minibatch size of 50, and was finally applied to produce the classification results on the provided test data.

As shown in Table 1, our 1-mix CNN-LTE system achieves an overall accuracy of 83.3% on the test data and outperforms the baseline by 6.1% absolute.

## 5. CONCLUSIONS

We presented our audio scene classification systems submitted for Task 1 of the DCASE 2016 challenge. These systems rely on LTE features automatically learned to encode the structure of the data. The systems based on individual LTE channels show comparable performance with the baseline whereas multi-channel LTE fusion leads to better accuracy. The best accuracies can be obtained by our 1-X pooling CNNs trained on multi-channel LTE images. Absolute improvement of 8.7% and 6.1% against the baseline is achievable with the 1-mix pooling CNN on the development and test data.

## 6. ACKNOWLEDGEMENT

This work was supported by the Graduate School for Computing in Medicine and Life Sciences funded by Germany's Excellence Initiative [DFG GSC 235/1].

## 7. REFERENCES

[1] http://www.cs.tut.fi/sgn/arg/dcase2016/.

[2] D. Wang and G. J. Brown, *Computational Auditory Scene Analysis: Principles, Algorithms, and Applications.* Wiley-IEEE Press, 2006.

[3] R. F. Lyon, "Machine hearing: An emerging field," *IEEE Signal Processing Magazine*, vol. 27, no. 5, pp. 131–139, 2010.

[4] R. Radhakrishnan, A. Divakaran, and P. Smaragdis, "Audio analysis for surveillance applications," in *Proc. IEEE Workshop on Applications of Signal Processing to Audio and Acoustics (WASPAA)*, 2005, pp. 158–161.

[5] S. Chu, S. Narayanan, C.-C. J. Kuo, and M. J. Mataric, "Where am I? Scene recognition for mobile robots using audio features," in *Proc. IEEE International Conference on Multimedia and Expo (ICME)*, 2006, pp. 885–888.

[6] Y. Xu, W. J. Li, and K. K. Lee, *Intelligent Wearable Interfaces.* Hoboken, NJ: Wiley, 2008.

[7] A. J. Eronen, V. T. Peltonen, J. T. Tuomi, A. P. Klapuri, S. Fagerlund, T. Sorsa, G. Lorho, and J. Huopaniemi, "Audio-based context recognition," *IEEE Trans. Audio, Speech, and Language Processing*, vol. 14, no. 1, pp. 321–329, 2006.

[8] H. Phan, L. Hertel, M. Maass, P. Koch, and A. Mertins, "Label tree embeddings for acoustic scene classification," in *Proc. ACM Multimedia 2016*, Amsterdam, The Netherlands, October 2016.

[9] H. Phan, L. Hertel, M. Maass, and A. Mertins, "Robust audio event recognition with 1-max pooling convolutional neural networks," in *Proc. Interspeech*, 2016, (to appear).

[10] H. Phan, L. Hertel, M. Maass, R. Mazur, and A. Mertins, "Representing nonspeech audio signals through speech classification models," in *Proc. Interspeech*, 2015, pp. 3441–3445.

[11] ——, "Learning representations for nonspeech audio events through their similarities to speech patterns," *IEEE/ACM Trans. Audio, Speech, and Language Processing*, vol. 24, no. 4, pp. 807–822, April 2016.

[12] L. Breiman, "Random forest," *Machine Learning*, vol. 45, pp. 5–32, 2001.

[13] A. Y. Ng, M. I. Jordan, and Y. Weiss, "On spectral clustering: Analysis and an algorithm," in *Proc. NIPS*, 2001, pp. 849–856.

[14] D. P. W. Ellis. (2009) Gammatone-like spectrograms. [Online]. Available: http://www.ee.columbia.edu/~dpwe/resources/matlab/gammatonegram/

[15] H. Phan, M. Maass, R. Mazur, and A. Mertins, "Early event detection in audio streams," in *Proc. IEEE International Conference on Multimedia and Expo (ICME 2015)*, July 2015.

[16] H. Phan, M. Maaß, R. Mazur, and A. Mertins, "Random regression forests for acoustic event detection and classification," *IEEE/ACM Trans. on Audio, Speech, and Language Processing*, vol. 23, no. 1, pp. 20–31, 2015.

[17] I. Laptev, M. Marszałek, C. Schmid, and B. Rozenfeld, "Learning realistic human actions from movies," in *Proc CVPR*, 2008, pp. 1–8.

[18] D. Barchiesi, D. Giannoulis, D. Stowell, and M. Plumbley, "Acoustic scene classification: Classifying environments from the sounds they produce," *IEEE Signal Processing Magazine*, vol. 32, no. 3, pp. 16–34, 2015.

[19] T. Heittola, A. Mesaros, A. J. Eronen, and T. Virtanen, "Audio context recognition using audio event histogram," in *Proc. European Signal Processing Conference (EUSIPCO)*, 2010, pp. 1272–1276.

[20] R. Cai, L. Lu, and A. Hanjalic, "Co-clustering for auditory scene categorization," *IEEE Trans. Multimedia*, vol. 10, no. 4, pp. 596–606, 2008.

[21] S. Deng, J. Han, C. Zhang, T. Zheng, and G. Zheng, "Robust minimum statistics project coefficients feature for acoustic environment recognition," in *Proc. IEEE International Conference on Acoustics, Speech and Signal Processing (ICASSP)*, 2014, pp. 8232–8236.

[22] J. Ye, T. Kobayashi, M. Murakawa, and T. Higuchi, "Acoustic scene classification based on sound textures and events," in *Proc. ACM Multimedia*, 2015, pp. 1291–1294.

[23] R. Martin, "Noise power spectral density estimation based on optimal smoothing and minimum statistics," *IEEE Trans. on Speech and Audio Processing*, vol. 9, no. 5, pp. 504–512, 2001.

[24] X. Glorot, A. Bordes, and Y. Bengio, "Deep sparse rectifier neural networks," in *Proc. 14th International Conference on Artificial Intelligence and Statistics (AISTATS)*, 2011, pp. 315–323.

[25] Y. Kim, "Convolutional neural networks for sentence classification," in *Proc. EMNLP*, 2014, pp. 1746–1751.

[26] A. Severyn and A. Moschitti, "Twitter sentiment analysis with deep convolutional neural networks," in *Proc. SIGIR*, 2015, pp. 959–962.

[27] Y. L. Boureau, J. Ponce, and Y. LeCun, "A theoretical analysis of feature pooling in visual recognition," in *Proc. ICML*, 2010, pp. 111–118.

[28] N. Srivastava, G. Hinton, A. Krizhevsky, I. Sutskever, and R. Salakhutdinov, "Dropout: A simple way to prevent neural networks from overfitting," *Journal of Machine Learning Research (JMLR)*, vol. 15, pp. 1929–1958, 2014.

[29] D. P. Kingma and J. L. Ba, "Adam: a method for stochastic optimization," in *Proc. International Conference on Learning Representations (ICLR)*, 2015, pp. 1–13.

[30] A. Mesaros, T. Heittola, and T. Virtanen, "TUT database for acoustic scene classification and sound event detection," in *Proc. EUSIPCO 2016*, 2016.